\pdfoutput=1
\documentclass[11pt]{article}

\usepackage[]{ACL2023}

\usepackage{times}
\usepackage{latexsym}
\usepackage{algorithmic}  
\usepackage{amsmath}  
\usepackage{algorithm} 
\usepackage[algo2e]{algorithm2e} 
\usepackage[utf8]{inputenc} % allow utf-8 input
\usepackage[T1]{fontenc}    % use 8-bit T1 fonts
\usepackage{hyperref}       % hyperlinks
\usepackage{url}            % simple URL typesetting
\usepackage{booktabs}       % professional-quality tables
\usepackage{amsfonts}       % blackboard math symbols
\usepackage{nicefrac}       % compact symbols for 1/2, etc.
\usepackage{microtype}      % microtypography
\usepackage{xcolor}         % colors
\usepackage{makecell}
\usepackage{xcolor,colortbl}
\usepackage{pifont,dsfont}
\usepackage{wrapfig,lipsum,booktabs}
\usepackage[export]{adjustbox}% http://ctan.org/pkg/adjustbox
\usepackage{caption}

\usepackage[most]{tcolorbox}
\definecolor{block-gray}{gray}{0.8}
\newtcolorbox{myquote}{colframe=black,boxrule=1pt,
colback=white,grow to right by=12mm,grow to left by=5mm,
boxsep=0pt,breakable}

\newtcolorbox{inside_myquote}{boxrule=0pt,
colback=block-gray,grow to right by=1mm,grow to left by=1mm,
top=0pt,bottom=0pt}

\usepackage{times}
\usepackage{soul}
\usepackage{url}
\usepackage{amsmath}
\usepackage{amsthm}
\usepackage{booktabs}
\usepackage{algorithm}
\usepackage{algorithmic}
\usepackage{caption}
\usepackage{makecell} 
\usepackage{graphicx}
\usepackage{subcaption}

\usepackage{xcolor,colortbl}

\newtheorem{definition}{Definition}

\newcommand{\hide}[1]{}
\newcommand{\myModel}{\textsc{TransNet}}

\usepackage[T1]{fontenc}
\usepackage[utf8]{inputenc}

\usepackage{microtype}

\usepackage{inconsolata}

\title{TransNet: Transfer Knowledge for Few-shot Knowledge Graph Completion}

\author{ Lihui Liu$^\dagger$,  Zihao Wang, Dawei Zhou, Ruijie Wang, Yuchen Yan$^*$, Bo Xiong, Sihong He, Kai Shu, Hanghang Tong$^*$ \\
Wayne State University \\
Hong Kong University of Science and Technology \\
Virginia Tech \\
Amazon \\
Stanford \\
The University of Texas at Arlington \\
University of Illinois Urbana-Champaign \\
}

\begin{document}
\maketitle

\begin{abstract}
Knowledge graphs (KGs) are ubiquitous and widely used in various applications. However, most real-world knowledge graphs are incomplete, which significantly degrades their performance on downstream tasks. Additionally, the relationships in real-world knowledge graphs often follow a long-tail distribution, meaning that most relations are represented by only a few training triplets. To address these challenges, few-shot learning has been introduced. Few-shot KG completion aims to make accurate predictions for triplets involving novel relations when only a limited number of training triplets are available. Although many methods have been proposed, they typically learn each relation individually, overlooking the correlations between different tasks and the relevant information in previously trained tasks. In this paper, we propose a transfer learning-based few-shot KG completion method (\myModel). By learning the relationships between different tasks, \myModel\ effectively transfers knowledge from similar tasks to improve the current task's performance. Furthermore, by employing meta-learning, \myModel\ can generalize effectively to new, unseen relations. Extensive experiments on benchmark datasets demonstrate the superiority of \myModel\ over state-of-the-art methods.
Code and datasets will be released upon acceptance.

\end{abstract}

\section{Introduction}

Knowledge graphs (KGs) are collections of triples $(h, r, t)$, where nodes represent real-world entities, events, or objects, and edges denote the relationships between them. KGs have numerous applications, such as in question answering ~\cite{gRetriever,Wang_Lipka_Rossi_Siu_Zhang_Derr_2024}, fact-checking ~\cite{factkg}, and recommender systems ~\cite{xiting,recommender}. However, real-world KGs are often incomplete, missing many relationships. For instance, in Wikidata, only about 30\% of people have a defined "date of birth" property ~\cite{metaR}. Knowledge graph completion aims to address this issue by predicting missing information based on the existing structure of the graph. Yet, a significant challenge in this task is the long-tail problem ~\cite{HiRe}, where a large portion of relations are associated with only a few triples, making it difficult to accurately complete the graph.

To address the challenges in knowledge graph completion, many few-shot KG completion methods have been proposed. For instance, HiRe ~\cite{HiRe} introduces a contrastive learning-based approach for few-shot KG completion. These methods aim to predict the missing tail entity $t$ for a query triplet by learning from only $K$ reference triplets associated with the target relation $r$. In $K$-shot KG completion, given a target relation $r$ and $K$ reference triplets, the goal is to accurately predict the tail entity $t$ for each query triplet $(h, r, ?)$ by leveraging the knowledge learned from the reference set. The key to success in few-shot KG completion lies in learning a meta-representation for each relation that can generalize from a limited number of reference triplets to novel relations.

Despite their success, existing few-shot KG completion methods generally treat each relation as an independent task, overlooking the correlations between tasks. In real-world knowledge graphs, however, relations are often interrelated. For example, the relation "has birthplace" is closely linked to "nationality," and knowledge transfer from one can therefore enhance the learning of the other. Building on this insight, exploring the relationships between different few-shot tasks and effectively transferring useful knowledge from related tasks to the target task—while avoiding negative transfer from irrelevant ones—remains a significant challenge.

%However, existing transfer learning algorithms cannot be directly applied to few-shot knowledge graph completion, as they typically rely on the IID assumption, which assumes that all source and target samples are drawn independently. Unfortunately, this assumption does not hold in knowledge graphs, where data is inherently non-IID. Additionally, the limited training data in few-shot tasks presents further challenges. 

In this paper, we focus on learning how to transfer knowledge between tasks for few-shot knowledge graph completion. Inspired by the Weisfeiler-Lehman (WL) subtree test ~\cite{wltree}, we propose an edge-based graph neural network model to learn task representations. These representations are then leveraged by a group-based transfer learning network to enhance performance across individual tasks. For each task, a meta-learning approach is employed to quickly adapt the model to unseen few-shot learning scenarios.

Specifically, we make the following key contributions:
\begin{itemize}
    \item We investigate knowledge transfer across different tasks in few-shot settings. By utilizing a relational message-passing graph neural network and a knowledge merge module, our approach effectively facilitates knowledge sharing across tasks.
    
    %\item To improve the model's ability to generalize to unseen relations, we employ a meta-learning-based framework. This allows the model to quickly adapt to new, unseen tasks.

    \item We conduct extensive experiments to rigorously evaluate the effectiveness and efficiency of our proposed approach, demonstrating its superior performance.
\end{itemize}

%%%%%%%%%%%%%%%%%%%%%%%%%%%%%%%%%%%%%%%%%%%%%%%%%%%%%%%%%%%%%%%%%%%%%%%
\section{Preliminaries and Problem Definition}\label{problem-definition}

A KG can be denoted as $\mathcal{G}=(\mathcal{V}, \mathcal{R}, \mathcal{L})$ where $\mathcal{V} = \{v_1, v_2, ..., v_n\}$ is the set of nodes/entities, $\mathcal{R} = \{r_1, r_2, ..., r_m\}$ is the set of relations and $\mathcal{L}$ is the list of triples.
Each triple in the KG can be denoted as $(h, r, t)$ where $h \in \mathcal{V}$ is the head (i.e., subject) of the triple, $t \in \mathcal{V}$ is the tail (i.e., object) of the triple and $r \in \mathcal{R}$ is the edge (i.e., relation, predicate) of the triple which connects the head $h$ to the tail $t$. 
The embedding of a node or relation type is represented by bold lowercase letters, e.g., $\mathbf{e}_{i}$, $\mathbf{r}_{i}$. 

In knowledge graph, the relations follow the long tail distribution. Most relation only have few training triplets. Few-shot knowledge graph link prediction focuses on learning a knowledge graph completion model with only a few samples.

\begin{definition}(Few-shot link prediction task T)
With a knowledge graph $\mathcal{G}=(\mathcal{V}, \mathcal{R}, \mathcal{L})$, given a support set $S_r = \{(h_i, t_i) \in \mathcal{V} \times \mathcal{V} | (h_i, r, t_i) \in \mathcal{L}\}$ about relation $r \in R$, where $|S_r| = K$, predicting the tail entity linked with relation $r$ to head entity $h_j$, formulated as $r : (h_j, ?)$, is called K-shot link prediction.
\end{definition}

As defined above, a few-shot link prediction task is always defined for a specific relation. During prediction, there usually is more than one triple to be predicted, and with support set $S_r$, we call the set of all triples to be predicted as query set $Q_r = \{r : (h_j, ?)\}$.

The objective of few-shot link prediction is to develop the ability to predict new triples for a relation $r$ after observing only a few example triples for $r$ (referred to as the support set). During training, the model is trained on a collection of tasks $T_{\text{train}} = \{T_i\}_{i=1}^M$, where each task $T_i = \{S_i, Q_i\}$ represents an individual few-shot link prediction scenario with its own support and query sets. Testing is performed on a different set of tasks $T_{\text{test}} = \{T_j\}_{j=1}^N$, which are similar to those in $T_{\text{train}}$, except that the tasks in $T_{\text{test}}$ involve relations not seen during training.

% \subsection{Transfer Learning}

% \subsection{Weisfeiler-Lehman (WL) subtree}

%%%%%%%%%%%%%%%%%%%%%%%%%%%%%%%%%%%%%%%%%%%%%%%%%%%

\section{Proposed Method}\label{overview}

%In this section, we investigate few-shot learning within the framework of knowledge graphs and provide a comprehensive overview of our proposed learning framework. 

\begin{figure*}[hbt!]
	\centering
	\includegraphics[width=0.84\textwidth]{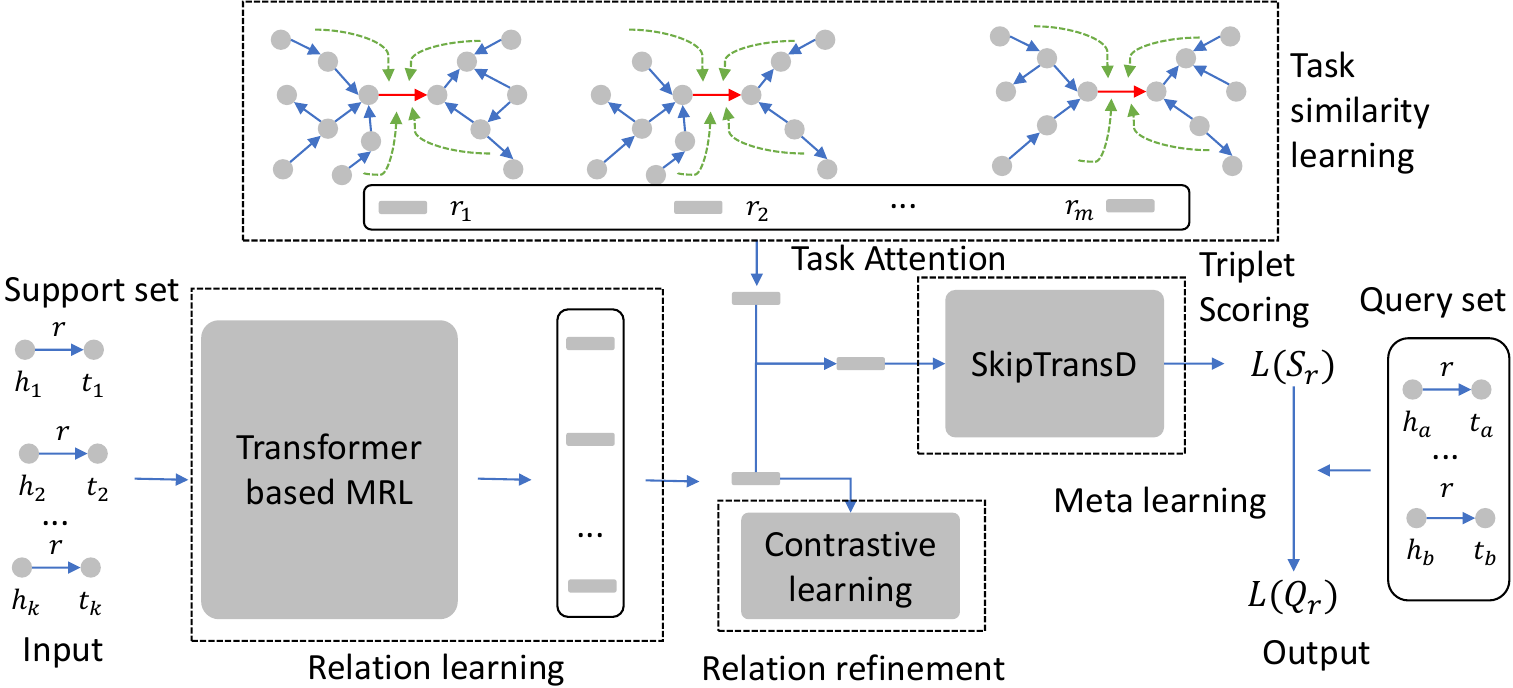}
	\caption{Architecture of our proposed \myModel\ Model. 
}
\label{fram}
	\vspace{-1\baselineskip}
\end{figure*}

In the context of $k$-shot knowledge graph completion, conventional methods tend to restrict learning to information contained within individual tasks. This fundamental limitation prevents the model from harnessing insights from related tasks, which in turn impairs its ability to develop more informative and generalizable representations for the target relation. To bridge this gap, we introduce an innovative approach that enables the transfer of valuable knowledge from related tasks to the target task, thereby enhancing performance in few-shot knowledge graph completion scenarios. The model architecture is shown in Figure ~\ref{fram}.

\subsection{Similarity measure between tasks}

In real-world applications, tasks often vary in similarity—some are closely related, making it beneficial to transfer knowledge between them, while others are unrelated or even negatively correlated, where transferring information can be harmful and degrade performance. 
%For example, as shown in Figure 2 \hh{Figure 2 is about contrastive loss. check}, the relations "hasHusband" and "hasWife" are closely connected and can support each other, whereas they do not share any meaningful correlation with the relation "isLocatedIn." 
The challenge in few-shot knowledge graph reasoning lies in effectively quantifying and further leveraging these correlations while avoiding the negative transfer of irrelevant information from unrelated tasks.

%Furthermore, most existing transfer learning algorithms rely on the IID (independent and identically distributed) assumption, which presumes that all source and target samples are drawn independently from the same distribution when assessing task similarity. However, this assumption does not hold in the non-IID nature of knowledge graphs. As a result, conventional transfer learning methods cannot be directly applied to the few-shot knowledge graph completion task.

To address this, we take the inspiration from the classic Weisfeiler-Lehman (WL) test ~\cite{wltree} and its intrinsic connection to the message-passing in graph neural networks (GNNs) ~\cite{hamilton2017inductive}. We propose using a relational message-passing GNN to encode task-specific information. The original WL test was designed to capture the local graph structure of a node. The structural information surrounding a node \( v \) is represented as a sequence of Weisfeiler-Lehman subtrees \(\{ f_0(v), f_1(v), \dots, f_m(v), \dots \}\) with increasing depth \( m \). Here, \( f_j \) is a relabeling function that encodes the subtree structure rooted at node \( v \), producing a new representation at each iteration. The similarity between two nodes is measured using their WL subtree similarity.

Unlike general graphs, whose nodes typically have features, in most knowledge graphs (KGs), edges (i.e., relations) have features, while nodes might not. This makes the node-based Weisfeiler-Lehman test not directly applicable to KGs. To overcome this limitation, we propose performing the WL test on edges instead of nodes. In this approach, the structural information surrounding an edge is represented as a sequence of Weisfeiler-Lehman edge subtrees, which we define as follows:

\begin{definition}(Weisfeiler-Lehman Edge Subtree)
    Given a relation \( r \), the Weisfeiler-Lehman edge subtree of depth \( m \) rooted at \( r \in \mathcal{R} \) is defined as:
    \[
    f_m(r) := f_m \left( f_{m-1}(r); \bigcup_{u \in N(r)} f_{m-1}(u) \right),
    \]
    where \( N(r) \) denotes the set of neighbors of the root edge \( r \).
\end{definition}

The Weisfeiler-Lehman (WL) edge subtree method aims to recursively encode the \(k\)-hop neighborhood information around an edge. If two edges share similar local subgraph structures, their corresponding WL edge subtrees will also be similar. 

When comparing the similarity between two few-shot tasks, each task can be represented as a set of relations or a sample of size \(K\) drawn from a specific distribution of WL edge subtrees. Therefore, measuring the similarity between two tasks is equivalent to comparing their respective sets of Weisfeiler-Lehman edge subtrees.
A commonly used approach to measure similarity between two sets of WL subtrees from different graphs is the Weisfeiler-Lehman Subtree Kernel which quantifies the discrepancy between two few-shot learning tasks by comparing their sets of WL edge subtrees.

\begin{definition}(Weisfeiler-Lehman Subtree Set Kernel)
Given two sets of subtrees, \(S\) with \(n\) subtrees and \(S'\) with \(n'\) subtrees, the Weisfeiler-Lehman subtree set kernel with \(M\) iterations is defined as:
\[
k(S, S') = \frac{1}{nn'} \sum_{m=0}^M \sum_{r \in S} \sum_{r' \in S'} \delta(f_m(r), f_m(r')),
\]
where \(\delta(\cdot, \cdot)\) is the Dirac kernel, which equals 1 when its arguments are identical and 0 otherwise. Here, \(f_m(r)\) represents the subtree pattern of depth \(m\) rooted at edge \(r\).
\end{definition}

%Following the Weisfeiler-Lehman (WL) edge subtree kernel, we assume that given a knowledge graph, the WL edge subtrees (i.e., \(\{ f_m(r) \mid r \in \mathcal{R} \}\)) at a fixed depth \(m\) are conditionally independent with respect to the WL edge subgraph \(G_{m-1}\) at depth \(m - 1\), i.e., \(f_m(r_1) \perp f_m(r_2) \mid G_{m-1}\). This assumption allows us to transform the non-IID transfer learning problem into an IID transfer learning problem. Specifically, the distribution shift between the edge subtrees of the source and target tasks can be measured by comparing the distribution shifts of their respective edge subtree structures.

To further extend the WL edge subtree kernel to handle continuous attributes and support edge aggregation, we draw inspiration from the connection between the WL edge subtree kernel and relational message-passing graph neural networks (GNNs). We propose measuring task similarity in the latent feature space induced by relational message-passing GNNs.
Combining these ideas, we define the discrepancy between two tasks as follows:

\begin{definition}(Task Discrepancy)
Given two sets of edge subtrees \(S\) with \(n\) subtrees and \(S'\) with \(n'\) subtrees, the task discrepancy between these edge subtree sets over \(M\) iterations is defined as:
\[
D(S, S') = \frac{1}{nn'} \sum_{m=0}^M \sum_{r \in S} \sum_{r' \in S'} \Theta(f_m(r), f_m(r')),
\]
where \(\Theta(\cdot, \cdot)\) is a neural network that measures the similarity between \(f_m(r)\) and \(f_m(r')\), and \(f_m(\cdot)\) represents the relational message-passing at depth \(m\) rooted at edge \(r\).
\end{definition}

The relational message-passing GNN is defined as:
\begin{equation*}
    m_i^r = \sum_{r' \in \mathcal{N}(r)} A(s_i^r, s_i^{r'})
\end{equation*}

\begin{equation*}
    s_{i+1}^r = U(s_i^r, m_i^r)
\end{equation*}
where \(\mathcal{N}(r)\) denotes the set of neighboring edges of \(r\) (i.e., edges that share at least one common endpoint with \(r\)) in the graph, and \(s_0^r\) is the initial edge feature of \(r\), representing the relation type.
And the similarity between two tasks is calculated as follows:
\begin{equation}\label{kernel}
k(T_1, T_2) = \sigma(\frac{1}{n n'} \sum_{m=0}^{M} \sum_{r \in S} \sum_{r' \in S'} f_m(r)^T W f_m(r')),
\end{equation}
where \(f_m(r)\) is the relational message-passing GNN at depth \(m\) rooted at edge \(r\), and \(\sigma(\cdot)\) is an activation function that measures the similarity between the representations. \(W\) is a learnable weight matrix that scales the similarity score.
During training,  to reduce the computational burden of Equation~\ref{kernel}, we approximate it by incorporating a task attention module.

\subsection{Relational learning}

In the previous section, we discussed how to measure the discrepancy between two tasks and how to learn the representation of a single relation. The next step is to learn a meta-representation for the target relation \(r\). To effectively learn this meta-representation, the model must satisfy two key properties. First, it should be insensitive to the size of the reference set (i.e., the few-shot size \(K\)). Second, the triplets in the reference set should be permutation-invariant, meaning that the order of the triplets should not impact the learning process. Additionally, the model needs to handle complex interactions between reference triplets to capture generalizable meta-relational knowledge. A transformer-based model is well-suited for this task. Transformers naturally satisfy the permutation-invariance property, as they do not assume any fixed ordering of the input data ~\cite{HiRe}. Furthermore, the attention mechanism in transformers allows the model to assign different weights to reference triplets, giving more importance to the triplets that are more representative of the relation. This makes transformers ideal for modeling pairwise interactions among reference triplets, ensuring that the learned meta-representations are both generalizable and robust. Therefore, we propose a transformer-based Meta Relation Learner (MRL) that captures pairwise triplet-triplet interactions within the support set \(S_r\), enabling the model to effectively learn generalizable meta-representations of the target relation.

Mathematically, given a meta-training task \(T_r\) targeting relation \(r\), our proposed Meta Relation Learner (MRL) takes the head/tail entity pairs from the reference set as input, i.e., \(\{(h_i, t_i) \in S_r\}\). Each reference triplet is encoded as \(x_i = h_i \oplus R_i \oplus t_i\), where \(h_i \in \mathbb{R}^d\) and \(t_i \in \mathbb{R}^d\) are the embeddings of the head entity \(h_i\) and the tail entity \(t_i\), and \(R_i \in \mathbb{R}^d\) is the embedding of the relation \(r_i\), all in dimension \(d\).

For all reference triplets associated with the same relation \(r\), the goal of the MRL is to capture the commonality among these reference (or support) triplets and generate a meta-representation for the target relation \(r\). This is achieved by encoding all the triplets with a transformer model, followed by a multi-layer perceptron (MLP), and then averaging the results across the reference set. The resulting meta-representation \(R\) for relation \(r\) is computed as follows:

\[
R = \frac{1}{K} \sum_{i=1}^K \text{MLP}(\text{transformer}(x_0, x_1, \dots, x_Q))
\]

To further incorporate task transfer learning into this framework, we propose the following mechanism to update the relation embedding based on neighboring relations. The updated embedding is given by:

\[
R_i' = \sigma \left( \sum_{j \in \mathcal{N}(i)} \alpha_{ij} W R_j \right)
\]

\noindent where \(\mathcal{N}(i)\) represents the set of neighboring relations, and \(\alpha_{ij}\) is the task similarity score between tasks \(T_i\) and \(T_j\), defined in the previous section as \(\alpha_{ij} = k(T_i, T_j)\), 
which is learned through an attention mechanism.
\(W\) is a learnable weight matrix, and \(\sigma\) is a non-linear activation function.

Finally, the relation representation \(R\) is updated as follows to combine the original meta-representation with the aggregated information from neighboring relations:

\[
R = \text{MLP}(R \, || \, R')
\]

Here, \(||\) represents the concatenation operation, and the MLP further refines the final relation representation.

\subsection{Triplet Scoring Module}

Ensuring the generalizability of the learned meta representation is essential for effective few-shot knowledge graph completion. While existing translation-based methods have proven successful, they are constrained by their reliance on Euclidean space. To address this limitation, it is critical to project entities into relation-specific spaces, where translational properties can be better learned. Drawing inspiration from skip connections, which ensure that new representations build upon and improve previous ones, we introduce \textbf{SkipTransD}. This model leverages the power of skip connections to enhance the learning of relational transitions in these customized spaces, offering a more flexible and powerful approach.
More specifically, given a triplet \((h_i, r, t_i)\), \textbf{SkipTransD} projects the head and tail entity embeddings into a latent space determined by the entities and the relation simultaneously, and at the same time, keeps the original embedding information. This process is mathematically expressed as:

\begin{equation*}
\begin{aligned}
    h_{p_i} &= \textrm{MLP}(r_{p_i}^\top h_{p_i} h_i) + h_i \\
    t_{p_i} &= \textrm{MLP}(r_{p_i}^\top t_{p_i} t_i) + t_i
\end{aligned}
\end{equation*}

The score function for each entity pair \((h_i, t_i)\) is then calculated as:

\begin{equation*}
    \text{l}(h_i, t_i) = \left\| h_{p_i} + R - t_{p_i} \right\|_2^2,
\end{equation*}

\noindent where \(\|x\|_2\) denotes the \(\ell_2\)-norm of vector \(x\), \(h_i\) and \(t_i\) represent the head and tail entity embeddings, \(h_{p_i}\) and \(t_{p_i}\) are their entity-specific projected vectors, and \(r_{p_i}\) is the relation-specific projected vector for relation \(r\). Finally, the loss function is defined as:
\[
L(S_r) = \sum_{(h_i, t_i) \in S_r} \max\{0, \text{l}(h_i, t_i) + \gamma - \text{l}(h_i, t'_i)\}
\]
where $\gamma$ is a hyper-parameter that determines the margin to separate positive pairs from negative pairs. $\text{l}(h_i, t'_i)$ calculates the score of a negative pair $(h_i, t'_i)$, which results from negative sampling of the positive pair $(h_i, t_i) \in S_r$, i.e., $(h_i, r, t'_i) \notin G$.

\section{Task-Conditioned Meta-Learning}

To enhance generalization and allow the model to quickly adapt to unseen relations, we propose a \textit{task-conditioned meta-learning strategy}. Unlike standard MAML ~\cite{mamal}, which applies a uniform gradient update, our approach modulates the adaptation step based on task-specific characteristics, ensuring better alignment between the learned representations and unseen relations.
Given a support set \( S_r \), we compute an \textit{adaptive scaling factor} \( \alpha_r \) that dynamically adjusts the update step based on the relation’s difficulty. Instead of using a fixed learning rate, we introduce
$\alpha_r = \sigma(W_r^\top \psi(S_r))$, where
\( W_r \) is a learnable transformation matrix,
\( \psi(S_r) \) extracts task-specific statistics (e.g., entity distribution, embedding variance),
\( \sigma(\cdot) \) is a sigmoid function to ensure \( \alpha_r \) stays between 0 and 1.
Using this, the relation representation is updated as:

\[
R_{T'}^r = R_r - \alpha_r \eta_r \nabla_{R_r} \mathcal{L}(S_r).
\]

This ensures that easier relations receive smaller updates, while more challenging relations get stronger adaptation.
We apply the same \textit{adaptive scaling} to entity updates:

\begin{equation*}
    \begin{aligned}
        h_{p_i}' &= h_{p_i} - \alpha_r \eta_r \nabla_{h_{p_i}} \mathcal{L}(S_r), \\
        r_{p_i}' &= r_{p_i} - \alpha_r \eta_r \nabla_{r_{p_i}} \mathcal{L}(S_r), \\
        t_{p_i}' &= t_{p_i} - \alpha_r \eta_r \nabla_{t_{p_i}} \mathcal{L}(S_r).
    \end{aligned}
\end{equation*}

This differentiates the adaptation process across tasks, making it more flexible and robust.
After adaptation, entity pairs \( (h_j, t_j) \) from the query set are scored using:

\[
\text{l}(h_j, t_j) = \left\| h_{p_j} + R_{T'}^r - t_{p_j} \right\|_2^2.
\]

and the final contrastive loss is defined as:

\[
\mathcal{L}(Q_r) = \sum_{(h_j, t_j) \in Q_r} \max\{0, \text{l}(h_j, t_j) + \gamma - \text{l}(h_j, t'_j)\},
\]
where \((h_j, t'_j)\) is a negative triplet sampled for contrastive learning. This training strategy ensures the model generalizes well across tasks and avoids overfitting to specific tasks, enabling better adaptation to unseen relations.

%%%%%%%%%%%%%%%%%%%%%

\subsection{Contrasive learning based relation refinement}

To further improve the relation representation quality, following ~\cite{HiRe}, a contrastive learning based method is used to make the representation more distinguishable. 

Given a target triplet \((h, r, t)\), its local content information is defined as: $C(h,r,t) = N_h \cup N_t$, 
where $N_h = \{(r_i, t_i) \mid (h, r_i, t_i) \in T_P\}$ and $N_t = \{(r_j, t_j) \mid (t, r_j, t_j) \in T_P\}$. 
Each relation-entity tuple \((r_i, t_i) \in C(h,r,t)\) is encoded as \(re_i = r_i || t_i\), where \(r_i \in \mathbb{R}^d\) and \(t_i \in \mathbb{R}^d\) are the embeddings of the relation and the entity, and \(||\) represents the concatenation of two vectors. 

To capture the local context information, a Multi-Head Self-Attention mechanism is applied. The context embedding \(c\) is computed as follows:
\[
c_0 = [re_1; re_2; \dots; re_K], \quad K = |C(h,r,t)|
\]
\[
c = \sum_{i=0}^{K} \alpha_i \cdot re_i, \quad \alpha = \text{MSA}(c_0)
\]
where \(c_0\) is the concatenation of all the embeddings from the context, and \(\alpha_i\) are the attention scores learned by the self-attention mechanism, allowing the model to weigh more relevant relation-entity pairs. %more heavily.

Furthermore, false contexts \(\{\tilde{C}(h,r,t)_i\}\) are generated by corrupting the relation or entity in each tuple \((r_i, t_i) \in C(h,r,t)\). The embeddings of these false contexts, \(\tilde{c}_i\), are learned using the same context encoder. To differentiate the true context from the false ones, a contrastive loss function is applied:
\[
\mathcal{L}_c = - \log \frac{\exp\left(\text{sim}(h || t, c) / \tau \right)}{\sum_{i=0}^{N} \exp\left(\text{sim}(h || t, \tilde{c}_i) / \tau \right)},
\]
where \(N\) is the number of false contexts, \(\tau\) is a temperature parameter, and \(\text{sim}(x, y)\) is the cosine similarity between vectors \(x\) and \(y\).

\begin{figure}[hbt!]
	\centering
	\includegraphics[width=0.4\textwidth]{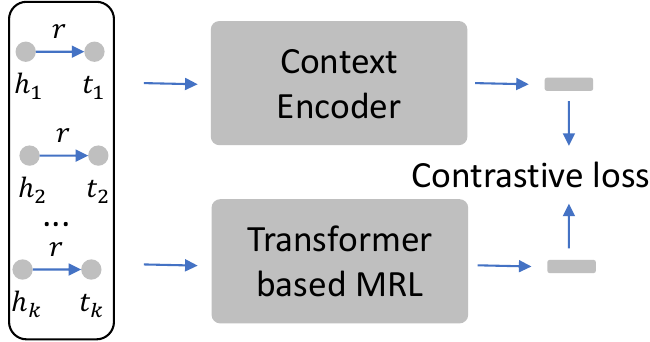}
	\caption{  
	The shared embedding space contains pre-trained embeddings of entities and relations by the  training data of KGC. Best viewed in color.
}
	\vspace{-1\baselineskip}
\end{figure}

\subsection{Model Warmup Phase}

When applying transfer learning to leverage knowledge from existing tasks, an excessive transfer can introduce noise, particularly when the target task is highly diverse. This often results in a decline in model performance. We observed this effect on the Wiki-One dataset, which contains over 180 tasks. A more effective approach is to first warm up the model without applying transfer learning. Once the model's performance stabilizes, transfer learning can be introduced gradually, followed by fine-tuning to adapt the model without overwhelming it with irrelevant information.  

\label{sec:method}

\section{Experiments }\label{experiments}
%\lh{Hi Professor, you can ignore this section. I am still working on it. I will let you know when I am done.}

%In this section, we conduct experiments to answer the following questions: (1) How accurate is the proposed \myModel\ algorithm for few-shot knowledge graph completion? (2) How effective are the proposed transfer learning and SkipTransD methods? The code and datasets will be made publically available upon the acceptance of the paper. 

%We evaluate \myModel\ for few-shot knowledge graph completion and assess the effectiveness of transfer learning. 

\begin{table*}[ht]
\centering
\caption{Performance on NELL-One and Wiki-One}
\label{tab:res}
\setlength{\tabcolsep}{1pt} % Reduces space between columns
\small % Change to \footnotesize or \scriptsize for smaller font size
\begin{tabular}{l|cc|cc|cc|cc|cc|cc|cc|cc}
\hline
 & \multicolumn{8}{c|}{Nell-One} & \multicolumn{8}{c}{Wiki-One}  \\ \hline
Methods & \multicolumn{2}{c|}{MRR} & \multicolumn{2}{c|}{Hits@10} & \multicolumn{2}{c|}{Hits@5} & \multicolumn{2}{c|}{Hits@1} & \multicolumn{2}{c|}{MRR} & \multicolumn{2}{c|}{Hits@10} & \multicolumn{2}{c|}{Hits@5} & \multicolumn{2}{c}{Hits@1} \\ \hline
         & 1-shot & 5-shot & 1-shot & 5-shot & 1-shot & 5-shot & 1-shot & 5-shot & 1-shot & 5-shot & 1-shot & 5-shot & 1-shot & 5-shot & 1-shot & 5-shot \\ \hline
TransE   & 0.105  & 0.168  & 0.226  & 0.345  & 0.111  & 0.186  & 0.041  & 0.082  & 0.036  & 0.052  & 0.059  & 0.090  & 0.024  & 0.057  & 0.011  & 0.042  \\
TransH   & 0.168  & 0.279  & 0.233  & 0.434  & 0.160  & 0.317  & 0.127  & 0.162  & 0.068  & 0.095  & 0.133  & 0.177  & 0.060  & 0.092  & 0.027  & 0.047  \\
DistMult & 0.165  & 0.214  & 0.285  & 0.319  & 0.174  & 0.246  & 0.106  & 0.140  & 0.046  & 0.077  & 0.087  & 0.134  & 0.034  & 0.078  & 0.014  & 0.035  \\
ComplEx  & 0.179  & 0.239  & 0.299  & 0.364  & 0.212  & 0.253  & 0.112  & 0.176  & 0.055  & 0.070  & 0.100  & 0.124  & 0.044  & 0.063  & 0.021  & 0.030  \\
ComplEx-N3 & 0.206 & 0.305  & 0.335  & 0.475  & 0.271  & 0.399  & 0.140  & 0.205  & OOM    & OOM    & OOM    & OOM    & OOM    & OOM    & OOM    & OOM    \\ \hline
GMatching & 0.185 & 0.201  & 0.313  & 0.311  & 0.260  & 0.264  & 0.119  & 0.143  & 0.200  & -      & 0.336  & -      & 0.272  & -      & 0.120  & -      \\
MetaR-I   & 0.250 & 0.261  & 0.401  & 0.437  & 0.336  & 0.350  & 0.170  & 0.168  & 0.193  & 0.221  & 0.280  & 0.302  & 0.233  & 0.264  & 0.152  & 0.178  \\
MetaR-P   & 0.164 & 0.209  & 0.331  & 0.355  & 0.238  & 0.280  & 0.093  & 0.141  & {0.314}  & 0.323  & 0.404  & 0.418  & 0.375  & 0.385  & 0.266  & 0.270  \\
FSRL      & -     & 0.184  & -      & 0.272  & -      & 0.234  & -      & 0.136  & -      & 0.158  & -      & 0.287  & -      & 0.206  & -      & 0.097  \\
FAAN      & -     & 0.279  & -      & 0.428  & -      & 0.364  & -      & 0.200  & -      & 0.341  & -      & 0.436  & -      & 0.395  & -      & 0.281  \\
HiRe      & 0.272 & 0.303 & 0.444 & 0.485 & 0.372 & 0.402 & 0.175 & \textbf{0.205}  & 0.296  & {0.343} & \textbf{0.425} & {0.453} & 0.393 & {0.412} & 0.220 & \textbf{0.289}  \\ \hline
 \myModel\ & \textbf{0.281} & \textbf{0.309} & \textbf{0.457} & \textbf{0.526} & \textbf{0.380} & \textbf{0.438} & \textbf{0.181} & {0.198}  & \textbf{0.306}   & \textbf{0.356}  & \textbf{0.425} & \textbf{0.472}  &  \textbf{0.394} & \textbf{0.439}  & \textbf{0.233}  & 0.285 \\ \hline
\end{tabular}
\vspace{-1\baselineskip}
\end{table*}

\subsection{Experiment Settings}

\textbf{Datasets.} In our experiments, we utilize two widely used few-shot knowledge graph (KG) completion datasets: NELL-One~\cite{HiRe} and Wiki-One~\cite{HiRe}. NELL-One is a subset of the NELL dataset, which is a system that continuously collects structured knowledge by reading the web and extracting facts from text. Wiki-One is a subset of Wikidata, which encompasses a much larger collection of entities and relationships. This dataset specifically emphasizes few-shot knowledge graph completion (KGC) within the context of large-scale knowledge graphs. 
Details of the dataset statistics are provided in Appendix.

\textbf{Baselines.} We evaluate \myModel\ against 11 baseline methods categorized into two groups: (1) Traditional KG completion methods, which include TransE~\cite{transE}, TransH~\cite{transH}, DistMult~\cite{distmult}, ComplEx~\cite{complEx}, and ComplEx-N3~\cite{complEx3N}. These models are trained using triplets from background relations, as well as relations from all reference sets and training relations; and (2) State-of-the-art few-shot KG completion methods, which include GMatching~\cite{GMatching}, MetaR~\cite{metaR}, FAAN~\cite{faan}, FSRL~\cite{fsrl}, and HiRe~\cite{HiRe}. For a fair comparison, we adopt the same settings as GMatching~\cite{GMatching}, selecting relations associated with more than 50 but fewer than 500 triplets for the few-shot completion tasks. For each target relation, we utilize the candidate entity set provided by GMatching~\cite{GMatching}.
Details of the model parameters are provided in Appendix.

\textbf{Evaluation Metrics.}
Following the common setting, We evaluate performance on both datasets using two common metrics: Mean Reciprocal Rank (MRR) and Hits@$n$ (with \(n = 1, 5, 10\)). MRR represents the average of the reciprocal ranks of the correct entities, while Hits@$n$ measures the proportion of correct entities that appear in the top-\(n\) ranked positions. We compare our proposed method to various baseline approaches under both 1-shot and 5-shot scenarios, as these are the most widely used settings in the prior work.

\subsection{Main Results}
In our experiment, \myModel\ was trained from scratch on the NELL-One dataset, while on the Wiki-One dataset, it underwent a warm-up phase without transfer learning before being fine-tuned with transfer learning.  
The performance metrics across the NELL-One and Wiki-One datasets reveal distinct strengths among various methods in few shot knowledge graph completion task. 
In particular, our proposed method, \myModel, achieves the highest Mean Reciprocal Rank (MRR) of 0.281 in the 1-shot setting on the NELL-One dataset, surpassing the next best method, HiRe, which scores 0.272. This improvement is also observed in the 5-shot setting, where \myModel\ reaches 0.309, indicating a consistent advantage over existing approaches. Notably, \myModel\ also leads in Hits@5 and Hits@1, showcasing its effectiveness in finding answers with high accuracy.

When examining the results on the Wiki-One dataset, \myModel\ maintains strong performance, 
achieving 0.439 in Hits@5 and 0.285 in Hits@1 in the 5-shot setting, 
further solidifying its status as a leading method for this task. 
The results indicate that \myModel\ effectively leverages the information from knowledge graphs to enhance prediction accuracy. Comparatively, traditional methods like TransE, TransH, and DistMult demonstrate lower performance across both datasets, underscoring the advantages of more advanced techniques such as \myModel. 
HiRe achieves the highest Hits@1 on the Wiki-One dataset in the 5-shot setting.

\subsection{Abalation Study}

\begin{table}[hbt!]
\centering
\caption{Ablation study of \myModel\ under 5-shot settings on Nell-One and Wiki-One
.}
\small
\label{table:ab}
\begin{tabular}{|l|c|c|c|c|c|}
\hline
\textbf{Dataset} & \multicolumn{4}{c|}{{Nell-One 5-shot}}   \\ \hline
 & {MRR} & {Hits@10} & Hits@5 & Hits@1  \\ \hline
\myModel\       & \textbf{0.309} & \textbf{0.526} & \textbf{0.438} & \textbf{0.198}   \\
-Transfer & 0.285 & 0.466 & 0.388 & 0.190 \\
-SkipTransD  & 0.279 & 0.465 & 0.386 & 0.175    \\ 
-Meta  & 0.243 & 0.428 & 0.359 & 0.151   \\ 
-Contrastive & 0.280 & 0.475 & 0.402 & 0.180 \\ \hline
\textbf{Dataset} & \multicolumn{4}{c|}{{Wiki-One 5-shot}}   \\ \hline
 & {MRR} & {Hits@10} & Hits@5 & Hits@1  \\ \hline
\myModel\       & \textbf{0.356} & \textbf{0.472} & \textbf{0.439} & \textbf{0.285}   \\
-Transfer & 0.328 & 0.435 & 0.405 & 0.265 \\
-SkipTransD  & 0.307 & 0.429 & 0.361 & 0.253    \\ 
-Meta  & 0.301 & 0.388 & 0.357 & 0.256   \\ 
-Contrastive & 0.318 & 0.454 & 0.395 & 0.263 \\ \hline
\end{tabular}
\vspace{-1\baselineskip}
\end{table}

\begin{table}[hbt!]
\centering
\caption{Comparison of Direct Training and Warm-Up on Wiki-One (5-shot and 1-shot)}
\small
\begin{tabular}{|l|c|c|c|c|}
\hline
\textbf{Dataset} & \multicolumn{4}{c|}{{Wiki-One 5-shot}}   \\ \hline
\textbf{Method} & \textbf{MRR} & \textbf{Hits@10} & \textbf{Hits@5} & \textbf{Hits@1} \\ \hline
Direct Training & 0.303 & 0.392 & 0.337 & 0.259 \\ 
Warm-Up         & 0.356 & 0.472 & 0.439 & 0.285 \\ \hline
\textbf{Dataset} & \multicolumn{4}{c|}{{Wiki-One 1-shot}}   \\ \hline
Direct Training & 0.229 & 0.367 & 0.315 & 0.148 \\ 
Warm-Up         & 0.306 &  0.425 & 0.394 & 0.233 \\ \hline
\end{tabular}
\vspace{-1\baselineskip}
\label{table:warm}
\end{table}

\noindent\textbf{Effectiveness of Transfer Learning.} The inclusion of transfer learning in \myModel\ significantly enhances its performance across various metrics. As shown in Table \ref{table:ab}, the model achieves a Mean Reciprocal Rank (MRR) of 0.309 with transfer learning, compared to 0.285 when transfer learning is omitted. This improvement underscores the importance of leveraging knowledge from related tasks, enabling the model to better generalize and predict missing relationships in the knowledge graph. Similar performance was observed on the Wiki-One dataset.

\noindent\textbf{Effectiveness of SkipTransD.} The results indicate that removing the SkipTransD method adversely affects the model's performance. The MRR drops to 0.279 without SkipTransD, which demonstrates its critical role in enhancing the predictive capabilities of \myModel. The reduction in Hits@10, Hits@5, and Hits@1 further illustrates that SkipTransD contributes significantly to accurately identifying relevant entities, thereby improving the overall effectiveness of knowledge graph completion.

\noindent\textbf{Effectiveness of Meta Learning.} The impact of excluding meta-learning is evident in the performance metrics, where \myModel\ achieves an MRR of only 0.243 without this component. This decline emphasizes the importance of meta-learning for adapting to new, unseen tasks quickly. By incorporating meta-learning, the model can efficiently leverage previous experiences, leading to improved accuracy in predicting tail entities across various contexts in the few-shot setting.

\noindent\textbf{Effectiveness of Warm-up.}
Table~\ref{table:warm} shows the performance of \myModel\ on the Wiki-One dataset. Note that warm-up is not applied to the NELL-One dataset because its tasks are less diverse.
The results indicate that warm-up helps reduce negative transfer when the task is highly diverse.

\vspace{-0.6\baselineskip}
\section{Related work}\label{related-work}
\vspace{-0.6\baselineskip}

\textbf{Knowledge Graph Completion.} Knowledge graph completion (KGC) aims to enrich knowledge graphs by predicting missing entities or relationships based on existing information. Traditional methods for KGC have primarily focused on embedding techniques, which represent entities and relations in a continuous vector space. Notable methods include TransE~\cite{transE}, which models relationships as translations in the embedding space, and DistMult~\cite{distmult}, which employs a bilinear approach to score triples. These methods have demonstrated the effectiveness of leveraging semantic information inherent in the graph structure to improve the accuracy of predictions. Additionally, more complex models such as ComplEx~\cite{complEx} and its variants have further enhanced KGC performance by capturing various types of interactions.

\textbf{Few-Shot Knowledge Graph Completion.} As knowledge graphs grow in scale and complexity, the challenge of completing them with limited labeled data has become increasingly important. Few-shot knowledge graph completion (FSKGC) addresses this challenge by leveraging techniques that enable models to learn from only a small number of examples. Recent advancements in FSKGC have employed meta-learning strategies, such as MetaR~\cite{metaR}, which utilizes relational patterns to generalize from a few instances, effectively improving performance in few-shot scenarios. Other approaches, like FAAN~\cite{faan} GANA ~\cite{gana} and GMatching~\cite{GMatching}, have explored novel architectures and training strategies to enhance the model's ability to transfer knowledge across tasks. These methods demonstrate the potential of few-shot learning to bridge the gap between the vastness of knowledge graphs and the scarcity of labeled data, enabling robust KGC even in resource-constrained environments. Other methods ~\cite{one,two,three,four} also solve this problem.

\textbf{Knowledge Graph Reasoning.} Generally speaking, there are two types of knowledge graph reasoning methods, including (1) embedding based approaches and (2) multi-hop approaches. For the former, 
the main idea is to learn a low dimensional vector for each entity and predicate in the embedding space, and use these embedding vectors as the input of the reasoning tasks (e.g.,~\cite{transE}, ~\cite{rotatE}, ~\cite{entity-predicate}, ~\cite{pprop}).
For the latter, the main idea is to learn missing rules from a set of relational paths sampled from the knowledge graph (e.g.,~\cite{williamCohen}, ~\cite{willianWang}, ~\cite{das-etal-2017-chains}). Many effective reasoning methods have been developed for predicting the missing relation (i.e., link prediction) or the missing entity (i.e., entity prediction). 
For example,
{\em TransR}~\cite{Lin2015TransR} learns the embedding of entities and predicates in two separate spaces. 
%The learned embedding (either by {\em TransE} or {\em TransR}) can be used for both link predication and entity predication. 
% %In ~\cite{SimplE}, \lliu{xxxxxxxxxx}. 
% In entity prediction, given the `subject' and the `predicate' of {\em a triple}, it predicts the missing `object'. For example, 
% %\hh{a few representative work, 1-2 sentences for each of them} 
% {\em GQEs}~\cite{entity-predicate} embeds the graph nodes in a low dimensional space, and treats the logical operators as learned geometric operations.

%%%%%%%%%%%%%%%%%%%%%%%%%%%%%%%%%%%%%%%%%%%%%%%%%%%%%%%%%%%%%%%%%%%%%%
\vspace{-0.6\baselineskip}
\section{Conclusion}\label{conclusion}
\vspace{-0.6\baselineskip}

In this paper, we address the critical challenge of knowledge graph completion in few-shot settings by proposing a novel approach that leverages knowledge transfer across interrelated tasks. By utilizing a relational message-passing graph neural network and an innovative knowledge merge module, our method enhances the ability to share information between tasks, thus improving the model's performance to unseen relations. The incorporation of a meta-learning framework further enables rapid adjustment to new tasks, demonstrating significant advancements in performance and efficiency through extensive experimental evaluation.

\section{Ethical Considerations}

We have thoroughly evaluated potential risks associated with our work and do not anticipate any significant issues. Our framework is intentionally designed to prioritize usability and ease of implementation, thereby reducing barriers for adoption and minimizing operational complexities. Furthermore, it's essential to highlight that our research builds upon an open-source dataset. This ensures transparency, fosters collaboration, and helps address ethical considerations by providing accessibility to the underlying data.

\section{Limitation}

Our dataset exhibits several limitations that warrant consideration. Firstly, our training data is constrained by its limited scope, primarily focusing on specific domains rather than providing comprehensive coverage across diverse topics. This restriction may affect the model's generalizability and performance in addressing queries outside of these predefined domains. Moreover, while our knowledge graph serves as a valuable resource for contextual information, it is essential to acknowledge its incompleteness. Despite its vast size, certain areas within the knowledge graph may lack sufficient data or connections, potentially leading to gaps in the model's understanding and inference capabilities.

\bibliography{custom}
\bibliographystyle{acl_natbib}

\clearpage
% \setcounter{secnumdepth}{0}
% \section{Supplementary Material: Reproducibility}\label{sec:appendix}

\appendix
\section{Appendix}
%\section{Appendix}

The statistics of both datasets are provided in Table ~\ref{dataset_statistics}. We use 51/5/11 and 133/16/34 tasks for training/validation/test on NELL-One and Wiki-One, respectively.

\begin{table}[h!]
\centering
\caption{Statistics of datasets. Each row contains the number of corresponding relations, entities, triplets, and tasks for each dataset.}
\setlength{\tabcolsep}{1pt}
\small
\begin{tabular}{|l|c|c|c|c|}
\hline
\textbf{Dataset} & \textbf{\#Relations} & \textbf{\#Entities} & \textbf{\#Triplets} & \textbf{\#Tasks} \\ \hline
NELL-One         & 358                  & 68,545              & 181,109             & 67               \\ \hline
Wiki-One         & 822                  & 4,838,244           & 5,859,240           & 183              \\ \hline
\end{tabular}
\label{dataset_statistics}
\end{table}

\subsection{Implementation Details}

To facilitate a fair evaluation, we initialize the \myModel\ model with entity and relation embeddings that were pretrained using TransE, as provided by GMatching. The embedding dimensions are chosen to be 100 for the NELL-One dataset and 50 for the Wiki-One dataset, in alignment with common practices in the field.

For training, we adopt a mini-batch gradient descent strategy with a batch size of 1,024 across both datasets. The Adam optimizer is utilized with a learning rate set to 0.001. Each reference triplet is associated with one negative context during training, and the trade-off parameter $\lambda$ is fixed at 0.05. To monitor performance, we evaluate the model on the validation dataset after every 1,000 steps, selecting the optimal model based on MRR over 30,000 steps. The entire implementation is carried out using PyTorch.

\end{document}